\documentclass[10pt,twocolumn,letterpaper]{article}

\usepackage[pagenumbers]{cvpr} % To force page numbers, e.g. for an arXiv version

\usepackage{multirow}
\usepackage{makecell}
\usepackage{gensymb}

\definecolor{cvprblue}{rgb}{0.21,0.49,0.74}
\usepackage[pagebackref,breaklinks,colorlinks,allcolors=cvprblue]{hyperref}

\def\paperID{16805} % *** Enter the Paper ID here
\def\confName{CVPR}
\def\confYear{2025}

\title{HyperPose: Hypernetwork-Infused Camera Pose Localization and an Extended Cambridge Landmarks Dataset}

\author{Ron Ferens \qquad Yosi Keller \\
Faculty of Engineering, Bar Ilan University, Ramat-Gan, Israel\\
{\tt\small \{ronferens, yosi.keller\}@gmail.com}
}
\begin{document}
\maketitle
% \input{sec/0_abstract}    
% \input{sec/1_intro}
% \input{sec/2_related_work}
% \input{sec/3_method}
% \input{sec/4_ecl_dataset}
% \input{sec/5_experimetns}
% \input{sec/6_ablation}
% \input{sec/7_conclusions}

%%%%%%%%%%%%%%%%%%%%%%%%%%%%%%%
%%% ABSTRACT
%%%%%%%%%%%%%%%%%%%%%%%%%%%%%%%
\begin{abstract}
In this work, we propose \textit{HyperPose}, which utilizes hypernetworks in absolute camera pose regressors. The inherent appearance variations in natural scenes, attributable to environmental conditions, perspective, and lighting, induce a significant domain disparity between the training and test datasets. This disparity degrades the precision of contemporary localization networks. To mitigate this, we advocate for incorporating hypernetworks into single-scene and multiscene camera pose regression models. During inference, the hypernetwork dynamically computes adaptive weights for the localization regression heads based on the particular input image, effectively narrowing the domain gap. Using indoor and outdoor datasets, we evaluate the HyperPose methodology across multiple established absolute pose regression architectures. We also introduce and share the Extended Cambridge Landmarks (ECL), a novel localization dataset, based on the Cambridge Landmarks dataset, showing it in multiple seasons with significantly varying appearance conditions. Our empirical experiments demonstrate that HyperPose yields notable performance enhancements for single- and multi-scene architectures. We have made our source code, pre-trained models \footnote{\scriptsize  \url{https://ronferens.github.io/hyperpose/}}, and the ECL dataset openly available \footnote{\scriptsize  \url{https://ronferens.github.io/extcambridgelandmarks/}}.
\end{abstract}

%%%%%%%%%%%%%%%%%%%%%%%%%%%%%%%
%%% INTRODUCTION
%%%%%%%%%%%%%%%%%%%%%%%%%%%%%%%
\begin{figure}[htbp]
    \centering
    \subfloat[Baseline architecture.]{
        \includegraphics[width=0.9\linewidth]{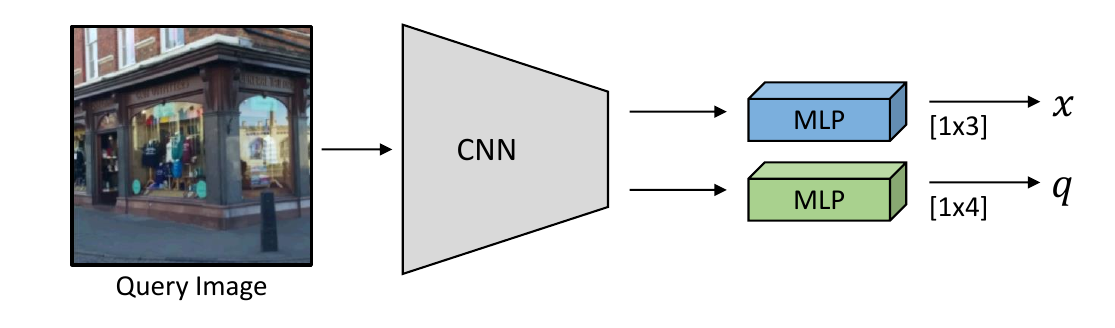}
        \label{fig:baseline_apr}
    }
    
    \subfloat[Baseline architecture with a hypernetwork.]{
        \includegraphics[width=0.9\linewidth]{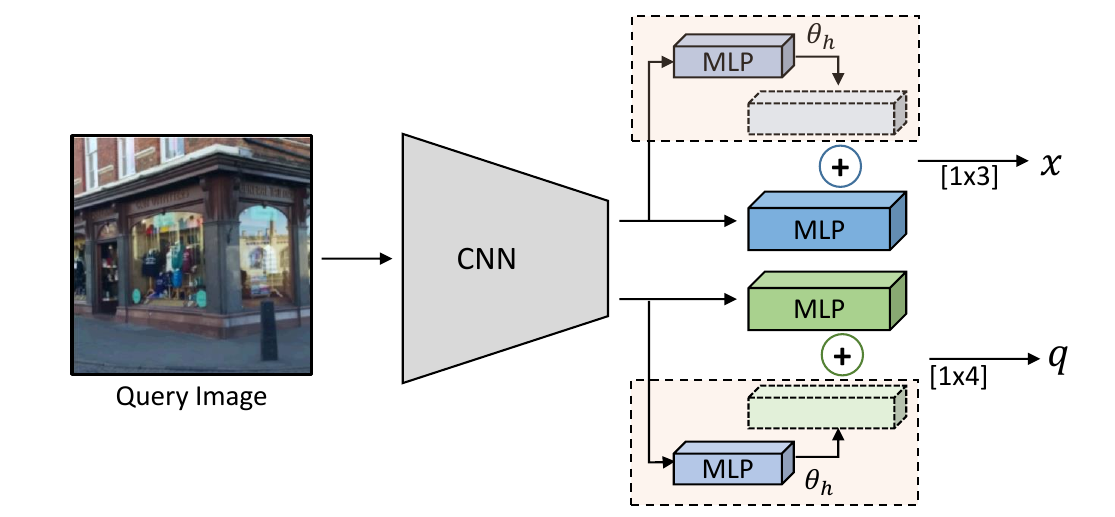}
        \label{fig:baseline_apr_w_hyper}
    }
    \caption{Figure~\ref{fig:baseline_apr} illustrates a baseline APR architecture. It consists of a CNN backbone that generates a feature vector encoding the input query image, followed by regression heads that estimate the translation and orientation. Figure~\ref{fig:baseline_apr_w_hyper} shows the extension of this baseline architecture with a hypernetwork that outputs the weights for the regression heads (translation and orientation) during inference.}
\label{fig:baseline_arch}
\vspace*{-1em}
\end{figure}
\vspace*{-1em}

\section{Introduction}
\label{sec:intro}
In many computer vision applications, such as indoor navigation, augmented reality, and autonomous driving, it is essential to determine the position and orientation of a camera using an image taken by that camera. Current camera localization techniques show various trade-offs between accuracy,
runtime, and memory usage. Feature Matching (FM) and Scene Coordinate Regression (SCR) are considered the most accurate localization methods. FM-based methods \cite{sattler2016efficient,taira2018inloc,sarlin2019coarse, sarlin2020superglue, sattler2015hyperpoints, tang2023neumap, li2020hierarchical, sarlin2021back, chen2024map} involve hierarchical
localization pipelines, which commonly use a coarse-to-fine processing
pipeline that begins with image retrieval to perform an initial localization based on images similar to the query image. The initial estimate is refined by local feature extraction and matching. To estimate the camera's 6-Degrees-Of-Freedom (6DoF) accurately, the resulting 2D-2D matches are mapped to 3D correspondences in the scene's point cloud and then used to determine the camera pose through Perspective-n-Point (PnP) and RANSAC. SCR-based methods \cite{DSAC, DSAC++, 9394752,
shotton2013scene, brachmann2019expert, 7780735} directly regress correspondences between 2D pixels and 3D coordinates in the scene.
Relative pose regression (RPR) schemes \cite{balntas2018relocnet,nn-net,en2018rpnet,ding2019camnet,turkoglu2021visual, chen2021widebaseline,winkelbauer2021learning} directly estimate the relative pose between a query image and a set of reference images with well-defined ground truth poses, eliminating the need for conventional pipelines.  However, an image retrieval scheme is essential to retrieve a set of nearest-neighbor images. Another approach, absolute pose
regressors (APRs), can estimate the camera pose with a single forward pass
using only the query image, resulting in significantly lower latency,
although notably less accurate. Additionally, due to their low memory
footprint, APRs can be deployed as a standalone application on edge devices
with limited computational resources. A survey of visual camera pose
localization is given by \cite{xu2022critical,shavit2019introduction}.

PoseNet by Kendall et al. \cite{kendall2015posenet} was the first APR approach, which leveraged a convolutional backbone and a multilayer perceptron (MLP) head to regress the camera's position and orientation. This simple and computationally efficient approach, enabling execution at 5ms, led to numerous APR methods that improved accuracy by modifying the backbone and MLP architectures \cite{melekhov2017image,naseer2017deep,walch2017image,wu2017delving,shavitferensirpnet,wang2020atloc,cai2019hybrid,song2024transbonet}. In addition, alternative loss functions and optimization strategies were explored \cite{kendall2016modelling,kendall2017geometric, shavit2019introduction}. Indoor and outdoor environments exhibit dynamic variability in illumination, perspective, and object motion. APR models, despite their computational efficiency, often struggle to maintain accuracy under these diverse conditions and show limited generalization to novel scenes, necessitating scene-specific training. Consequently, while architecturally simple, a unified network might yield suboptimal performance when adapting to heterogeneous real-world visual inputs. Recent studies have proposed extending APR from single-scene to multiscene formulations. Blanton et al. \cite{blanton2020extending} implemented a convolutional neural network (CNN) backbone to compute a global descriptor for the input image. Rather than employing a scene-specific multilayer perceptron (MLP), they optimized a set of fully connected (FC) layers, allocating one layer per scene based on a predicted scene identifier. Although introducing a framework to optimize a single model across multiple scenes, this methodology did not achieve state-of-the-art (SOTA) APR accuracy. An alternative approach was presented
by \cite{shavit2021learning} using dual-branch
transformers \cite{AttentionIsAllYouNeed}. Encoders focused on extracting
pose-related features, while decoders transformed encoded scene identifiers
into latent pose representations. The decoder outputs are leveraged for
scene classification and selection of corresponding position and orientation
embeddings, which are used to regress position and orientation vectors.

Hypernetworks \cite{ha2016hypernetworks} are a deep learning architecture in
which an auxiliary network, the hypernetwork, generates the weights for the
primary network (main network) based on the current input and downstream
task. The main network and hypernetwork are trained end-to-end using backpropagation. The hypernetwork allows for dynamic and adaptive weight modulation of the main network, leading to improved flexibility in response
to changing conditions \cite{nirkin2021hyperseg, chen2022transformers, dong2021robust, tay2021hypergrid}.

In this work, we propose \textit{HyperPose}, a hypernetwork-based approach for absolute camera pose regression. Extending common APR methods \cite{kendall2015posenet, kendall2016modelling, walch2017image,
shavitferensirpnet, cai2019hybrid, ShavitFerensIccv21}, the hypernetwork
computes weights for the regression head. This enables adaptive analysis of image content, allowing the regression weights to focus on informative
features, rather than relying on a single constant pretrained set of
weights. We validated the proposed method by integrating
hypernetworks into single- and multiscene APR architectures. For
single-scene networks, we present HyperPose with an existing state-of-the-art APR model \cite{wang2020atloc} as well as a simplistic baseline APR using an
off-the-shelf backbone \cite{tan2019efficientnet} and shallow regression
heads, inspired by the principles outlined in \cite{kendall2015posenet}. For multiscene, we extend a leading multiscene APR model \cite{shavit2021learning} by integrating a hypernetwork architecture. In both instances, hypernetwork-augmented models demonstrated superior performance compared to their original counterparts when evaluated on the benchmark datasets in this study.
To evaluate localization robustness, we introduce an Extended Cambridge Landmarks (ECL) dataset - an augmentation of the original benchmark \cite{kendall2015posenet} with algorithmically generated environmental shifts. By combining evening, winter, and summer scenes, the ECL incorporates dynamics such as lighting shifts and seasonal transitions.

Thus, we propose the following contributions:
\begin{itemize}
\item We propose an approach to enhance any existing absolute camera pose
regression models by integrating hypernetwork architectures.

\item We comprehensively evaluated the proposed HyperPose on
both single- and multiscene pose regressors, demonstrating improved accuracy
achieved by the hypernetwork-enhanced models over the original versions.

\item The proposed scheme achieves new SOTA accuracy on multiscene APR
benchmarks spanning diverse outdoor and indoor localization tasks.

\item We introduce the Extended Cambridge Landmarks (ECL) dataset, which builds upon the foundation of the original Cambridge Landmarks dataset \cite{kendall2015posenet}. Our ECL data set extends existing test scenes by incorporating various appearance conditions.
\end{itemize}

%%%%%%%%%%%%%%%%%%%%%%%%%%%%%%%
%%% RELATED WORK
%%%%%%%%%%%%%%%%%%%%%%%%%%%%%%%
\section{Related Work}
\label{sec:relatedwork}

%=========================================
% Camera Pose
%=========================================

Camera pose estimation techniques differ by their input during inference and
algorithmic attributes.

\textbf{3D-based or structure-based localization} methods are considered the most accurate for camera pose estimation and have demonstrated
state-of-the-art performance in leading benchmarks. These methods utilize
correspondences between 2D features in an image and 3D world coordinates.
\cite{sattler2016efficient,taira2018inloc,sarlin2019coarse, sarlin2020superglue, sattler2015hyperpoints, tang2023neumap, li2020hierarchical, sarlin2021back, chen2024map, radwan2018vlocnet++}
introduced a two-phase approach for hierarchical pose estimation pipelines.
Thus, each query image is first encoded using a CNN trained for
image retrieval (IR), based on a pre-mapped image dataset. Tentative
correspondences are then estimated by matching local image features to 3D
matches. The resulting matches are used by PnP-RANSAC to estimate the camera pose. DSAC \cite{DSAC} and its succeeding scheme, DSAC++ \cite{DSAC++}%
, utilize a CNN architecture to directly estimate 3D coordinates from 2D
positions in an image and only require the query image during inference. The
same as in Absolute Pose Regression (APR) methods, separate models must be
trained for each scene, achieving state-of-the-art accuracy. Lately, \cite{chen2024map} proposes a solution to mitigate this gap by introducing a scene-agnostic architecture.

\textbf{Relative Pose Regression} (RPR) algorithms estimate the relative
pose between an input query image and a set of reference images based on
their known location. Calculating the relative pose involves estimating the relative position and orientation between the query image and the reference images \cite{essnet, 9394752, arnold2022map}. An Image Retrieval (IR) approach is utilized to determine the closest set of neighbor images, and the relative
motion is then estimated between the query image and each of the retrieved
images. The estimated relative poses are used, along with the known
position of the reference images, to estimate the camera's absolute pose.
Similar to 3D and structured-based localization methods, RPRs involve a
multistep process and require access to a database of images with labeled
poses during the inference phase.

\textbf{Absolute Pose Regression} (APR) directly estimates the camera pose
given the input query image. APRs were first introduced by Kendall \cite%
{kendall2015posenet}, using a modified version of the truncated GoogLeNet
architecture, where the Softmax classification layer was replaced by a
series of fully connected layers that output the pose estimate. Although less accurate than classical structure-based methods, APRs offer several advantages, such as faster inference times
(milliseconds), reduced memory footprint (megabytes), and do not require
feature engineering. Variations of the encoder and MLP architectures have
been proposed to improve APR accuracy \cite%
{naseer2017deep,shavitferensirpnet,naseer2017deep,kendall2016modelling,cai2019hybrid, song2024transbonet}%
. Furthermore, modifications to the loss function have been suggested to
improve the balance between orientation and position objectives \cite%
{kendall2017geometric} or to incorporate other types of information \cite%
{brahmbhatt2018geometry}. For an in-depth examination of the various
architectural designs and loss functions utilized in camera pose regression,
please refer to \cite{shavit2019introduction}. Although demonstrating
faster inference times and low memory requirements, absolute pose
regression (APR) models typically require dedicated per-scene training to
achieve competitive accuracy compared to other methods. This limitation has
raised concerns regarding the value of APRs versus significantly more
accurate techniques such as \cite{DSAC, DSAC++, 9394752,
ramachandran2017searching}. To address this issue, recent approaches
suggested extending APRs to a multiscene paradigm. \cite%
{blanton2020extending} introduced Multi-Scene PoseNet (MSPN), which utilizes
a CNN to regress an image feature vector. The two
branches then process this feature vector. The first branch classifies the
related scene using a fully connected layer with a softmax operator. The
estimated scene index selects scene-specific weights from a database. The
second branch combines these weights with the feature vector to regress the
pose. The model is trained end-to-end by applying both binary cross-entropy
(for classification) and camera pose losses for each scene. Inspired by
recent successful applications of Transformers \cite{DETR, 16x16,liu2021swin}
to computer vision tasks, \cite{shavit2021learning} suggested a shared
convolutional network backbone followed by two network branches consisting
of a Transformer and a regression head. The first regresses the input
query image's pose, and the other its orientation. The main
incentive for utilizing Transformers encoders is to focus on
pose-informative features and decoders to transform encoded scene
identifiers into latent pose representations.

\begin{figure*}[tbp]
\centering
\includegraphics[width=\linewidth]{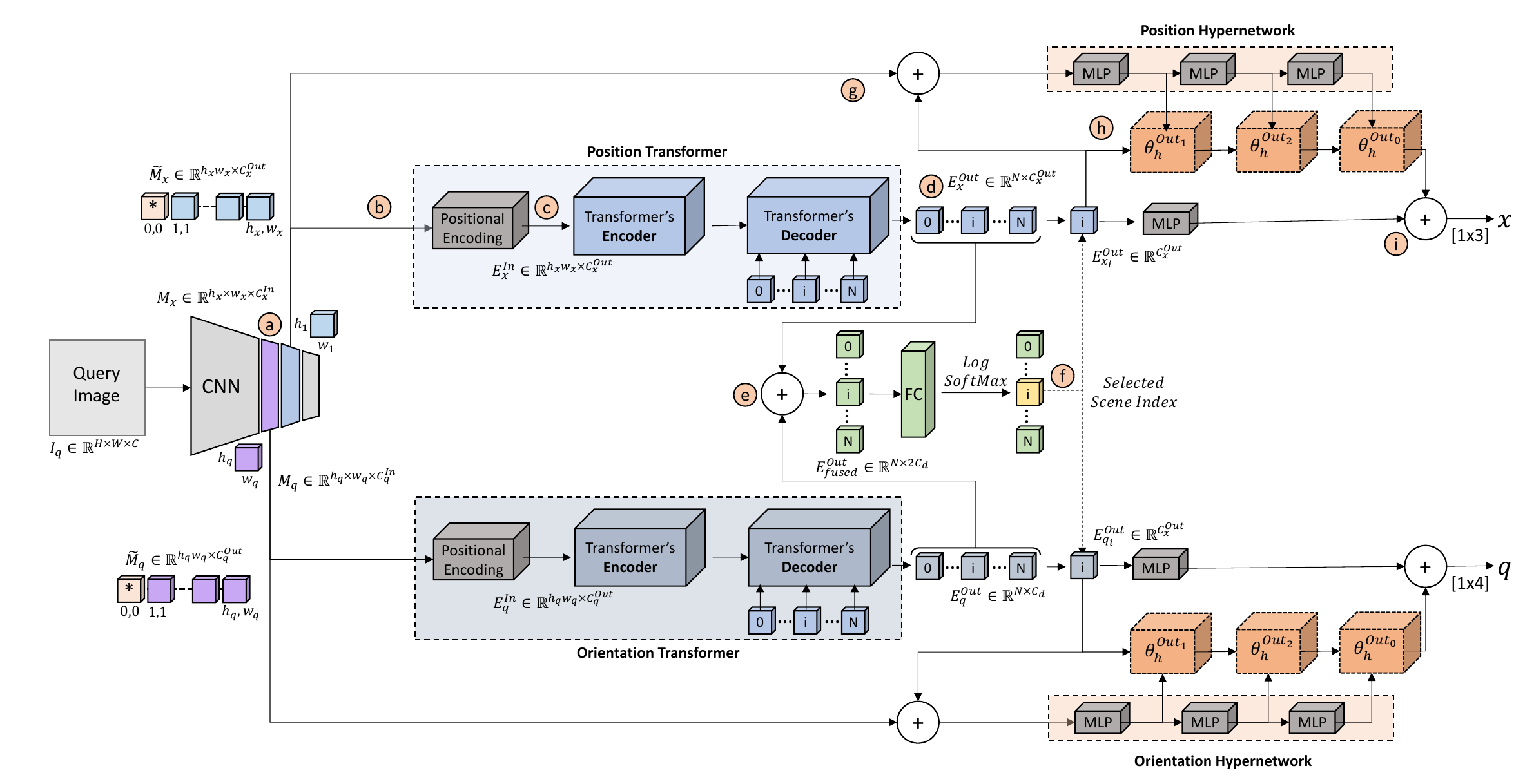}
\captionof{figure}{\textbf{MS-Transformer with hypernetwork (MS-HyperPose)} - The proposed multi-scene absolute pose regression architecture using a hypernetwork. The primary network employs position and orientation Transformers, in a dual-branch architecture, to extract activation maps from the underlying convolutional backbone. The hypernetwork generates the weights for the regression head in the primary network, using the input query image. These adaptive weights, combined with the latent vectors generated by the Transformers estimate the camera pose, consisting of the spatial position (x) and angular orientation (q).}
\label{fig:architecture}
\end{figure*}

%=========================================
% Hypernetworks
%=========================================

\textbf{Hypernetworks} were first presented by Ha et al. \cite{ha2016hypernetworks}.
These are neural networks that have been designed to predict the weights of
a primary network. The primary network's weights can be adjusted based on
specific inputs, resulting in a more expressive and adaptive model.
Hypernetworks have been used in various applications such as learning
implicit neural representations \cite{chen2022transformers}, semantic
segmentation \cite{nirkin2021hyperseg}, 3D scene representation and modeling
\cite{littwin2019deep, sitzmann2020implicit, spurek2022hyperpocket} and
continuous learning \cite{von2019continual}, to name a few.

%%%%%%%%%%%%%%%%%%%%%%%%%%%%%%%
%%% METHOD
%%%%%%%%%%%%%%%%%%%%%%%%%%%%%%%
\section{APRs using Hypernetworks}
\label{sec:method}

HyperPose is a general approach, applicable to both single- and multi-scene scenarios, for enhancing APR schemes. It comprises two key components: the main network (APR) designed to estimate the camera's absolute position and the hypernetwork that computes the regression weights for the main network conditioned on the input query image. The main network performs a forward pass on the input image to localize the capture camera. The camera pose is represented as the tuple $<\mathbf{x},\mathbf{q}>$, where $\mathbf{x}\in \mathbb{R}^{3}$ is the camera's position in world coordinates and $\mathbf{q} \in \mathbb{R}^{4}$ is the quaternion encoding of its $3D$ orientation. Following \cite{chai2020supervised}, we formulate the hypernetwork $H$ as a parametric function $\mathbf{\theta }{_{h}}=H(\mathbf{\theta },\mathbf{x})$, where $\mathbf{\theta }$ represents the parameters of the hypernetwork, $\mathbf{x}$ the network's input, and $\mathbf{\theta }{_{h}}$ the parameters of the regression heads layer for the input query image $I_{q}$.

%=========================================
% Network Arch
%=========================================
\subsection{Network Architecture}

\subsubsection{HyperPose for Single-Scene APRs}
\label{sec:single-scene-apr} We apply the proposed HyperPose-based approach to single-scene APR using a baseline APR following 
\cite{kendall2015posenet}, and the
state-of-the-art \textit{AtLoc} APR \cite{wang2020atloc}.

\textbf{Baseline APR with Hypernetwork.} As depicted in Fig. \ref{fig:baseline_apr}, the baseline model's main network consists of a CNN backbone that generates a feature vector encoding the input query image. This vector is then processed by two regression
heads. The first estimates the camera position, and the other
estimates the orientation. Figure \ref{fig:baseline_apr_w_hyper} illustrates the
integration of a hypernetwork into this architecture. The input to the hypernetwork is the feature vector generated by the CNN backbone of the main network. This vector is passed through a single fully-connected (FC) layer followed by a Swish activation \cite{ramachandran2017searching} function. The processed vector serves as an input
to a multilayer perceptron (MLP) that outputs the weights for a single
layer in the regression heads of the main network. The camera
pose is then estimated by the position and orientation regression heads of the main network.

\textbf{AtLoc with Hypernetwork.} \cite{wang2020atloc} presents Attention-Guided Camera Localization (AtLoc), an absolute pose regression model based on self-attention. In this
architecture, a visual encoder encodes an image into a latent vector. Conditioned on these extracted features, an attention module
computes self-attention maps to reweight the representation into a new
feature space. A pose regressor then maps the transformed features after the
attention operators to estimate the camera pose. We incorporate a hypernetwork into the AtLoc, the same as for the baseline APR model. In this case, the input to the hypernetwork is the reweighted  
representation feature vector computed by the visual encoder.

\subsubsection{HyperPose for Multi-Scene APRs} The suggested multi-scene model, named \textit{MS-HyperPose} is an extension of the architecture introduced in \cite{ShavitFerensIccv21}. As depicted in Fig. \ref{fig:architecture}, the input query image $I_{q}$ is processed by a convolutional backbone that computes intermediate activation maps from two different layers. These activation maps are then utilized as inputs to the main network and the hypernetwork. Within the main network, the
positional and orientational Transformers process the activation maps and
integrate sequential representations into single latent vectors. After
converting the activation maps into representation vectors, these vectors, along with the latent vectors are passed through multilayer perceptron (MLP)
layers in the hypernetwork to generate the weights $\mathbf{\theta }_{h}$
for the translation and orientation regression heads. Using the weights $%
\mathbf{\theta }_{h}$ produced by the hypernetwork $H$ and the regression
heads in the main network, the position, and orientation of the input query
image are estimated.

\textbf{The Main Network} consists of a convolutional backbone and two
separate branches to regress the camera's position and orientation. Each
branch includes its particular Transformer and a multilayer perceptron (MLP) head.
Studies have demonstrated that substituting the patchify stem, utilized in
the early visual processing of a Vision Transformer (ViT), with a convolutional stem enhances optimization stability and elevates performance to a higher level \cite{xiao2021early}. Therefore, given an input
query image $I_{q}\in \mathbb{R}^{H\times W\times C}$, a shared
convolutional backbone generates activation maps. Two different activation
maps are selected from the CNN based on the particular learned task (Fig. 2a). The
activation maps, represented as $M\in \mathbb{R}^{h\times w\times C^{In}}$, are processed and transformed into $\Tilde{M}\in \mathbb{R}^{h\times w\times C^{Out}}$ using a $1\times 1$ convolutional layer followed by flattening (Fig. 2b).

Following \cite{ShavitFerensIccv21}, we use the Transformer architecture from \cite{DETR}, modifying the standard Encoder and Decoder to add positional encoding at each attention layer. Each Transformer has 
$L$ identical blocks, each with a multi-head attention (MHA) mechanism and a two-layer MLP with Swish activation \cite{ramachandran2017searching}. Layer Normalization (LN) is applied before MHA and MLP processing, as in \cite{AttentionIsAllYouNeed}, with outputs merged via residual connections and dropout. To preserve spatial information, each activation map position receives a learned encoding. The processed sequence, serving as input to the Transformer is given by (Fig. 2c):

\vspace*{-1em}
\begin{equation}
E^{In}=[T,\Tilde{M}]\in \mathbb{R}^{(hw)\times
C^{Out}}, T\in \mathbb{R}^{C_{M}^{Out}}. 
\label{eq: main net encoder input}
\end{equation}

where $T$ is the positional encoding of the activation map.

Given a dataset with $N$ scenes, the Transformer Decoders generate embeddings $E^{Out}=\in \mathbb{R}^{N\times
C^{Out}}$ (Fig. 2d). For a query image, only one position matches its scene. The combined Transformer outputs (Fig. 2e) pass through a fully connected layer and Log SoftMax, selecting the highest-probability vectors (Fig. 2f).

The selected vectors $E_{x_{i}}^{Out}$ and $E_{q_{i}}^{Out}$ are processed through an MLP. Finally, the estimated camera position and orientation are obtained by summing the outputs of the main network and hypernetwork, using weights predicted by the hypernetwork (Fig. 2h and 2i).

\textbf{The Hypernetworks} consist of MLPs, each regressing the weights for a corresponding regression head (position or orientation) in the main network. Their input is the sum of task-specific encoded activation maps $\Tilde{M_{x}}$ and $\Tilde{M_{q}}$, along with global latent vectors $E_{x_{i}}^{Out}$ and $E_{q_{i}}^{Out}$ (Fig. 2g). To ensure size compatibility, the encoded activation map passes through a fully connected layer before summation. This vector is then processed by MLPs with Swish activation, generating inputs for another MLP set, each estimating weights for a specific layer in the main network’s regression heads:

\vspace*{-1em}
\begin{equation}
\theta _{h}^{{Out}_{j}}=H(\theta _{h}^{{Out}_{j}},E_{i}^{Out})\in \mathbb{R}^{{C_{E_{i}}^{Out}} \times C_{H}^{{Out}_{j}}}, 
\label{eq: hyper theta output}
\end{equation}

where $j$ is the index of the regression head layer.

%=========================================
% Optimization Criteria
%=========================================
\subsection{Optimization Criteria}

The localization error is calculated by comparing the translation position
and orientation of the ground truth pose $\mathbf{p}_{gt}=<\mathbf{x}_{gt},%
\mathbf{q}_{gt}>$, where $\mathbf{x}\in \mathbb{R}^{3}$ represents the
position of the camera in the world and $\mathbf{q}\in \mathbb{R}^{4}$
denotes its orientation encoded as a quaternion, with the estimated pose $%
\mathbf{p}_{est}=<\mathbf{x}_{est},\mathbf{q}_{est}>$. The translation
error, represented by $L_{x}$, is
calculated by the Euclidean distance between the ground truth and the
estimated camera positions. The orientation error, $L_{q}$, is defined as
the Euclidean distance between the ground truth and the estimated unit quaternions 
\cite{shavit2019introduction}:

\vspace*{-2em}
\begin{equation}
L_{x}=||\mathbf{x}_{0}-\mathbf{x}||_{2},  \label{eq: lx loc loss}
\end{equation}%
\begin{equation}
L_{q}=\left\Vert \mathbf{q_{0}}-\frac{\mathbf{q}}{||\mathbf{q}||}\right\Vert
_{2}  \label{eq: lq rot loss}
\end{equation}
where $\mathbf{q}$ is a normalized norm quaternion to ensure it is a valid
orientation encoding.

As a multi-regression optimization problem, we adopt the method outlined in
\cite{kendall2017geometric} by using different weighting schemes, based on
the uncertainty of each task, to balance the individual loss functions
defined for the separate objectives. The model optimizes the aggregated loss
function, a combination of the weighted individual losses.%

\vspace*{-2em}
\begin{equation}
L_{p}=L_{x}\exp (-s_{x})+s_{x}+L_{q}\exp (-s_{q})+s_{q},
\label{eq: total pose loss}
\end{equation}

where $s_{x}$ and $s_{q}$ are the learned parameters. As a further optimization, we adopt the approach detailed in \cite{shavitferensirpnet}
and follow a three-step training process. Firstly, the entire network is trained using the aggregated loss in Eq. \ref{eq: total pose loss}. In the next two stages, each MLP head undergoes separate fine-tuning with its corresponding loss function.

%=========================================
% Implementation Details
%=========================================
\subsection{Implementation Details}
The shared backbone uses
EfficientNet-B0 \cite{pmlr-v97-tan19a,efficientnet-pytorch}. We used the activation maps of two endpoints (reduction levels) as
input for our position and orientation branches: $m_{rdct4}\in \mathbb{R}%
^{14\times 14\times 112}$ and $m_{rdct3}\in \mathbb{R}^{28\times 28\times
40} $, respectively. Following \cite%
{ShavitFerensIccv21}, we applied linear projections on each activation map
to a common depth dimension of $C_{M}^{Out}=256$ and learned the positional
encoding of the same depth. The regression
heads in the main network consist of two-layer MLPs with input and output layers. The input layer dimension is 1024 for both the position and orientation, while the output dimensions are 3 and 4, respectively.

\textbf{Single-Scene APR.} For both the baseline APR and the HyperPose
extension of the AtLoc \cite{wang2020atloc} model, the hypernetwork for the
position branch generates weights for an MLP with a dimension of 256, while
the orientation hypernetwork produces weights of 512 dimensions.
\begin{table*}[tbh]
\caption{\textbf{Comparative analysis of APR architectures with and without the purposed hypernetwork addition using the Cambridge Landmarks dataset}. We report the median position/orientation error in meters/degrees. The best performance is marked in \textbf{bold}. (*) The authors of AtLoc did not report results on the Cambridge dataset, and we trained the AtLoc model on this dataset.}
\label{tb:singlescene_comapre_cambridge}
\centering
\small
\begin{tabular}{cccccc}
\hline
\textbf{Method} & \textbf{K. College} & \textbf{Old Hospital} & \textbf{Shop Facade} & \textbf{St. Mary} & \textbf{Avg.} \\ \hline
\multicolumn{1}{l}{AtLoc* \cite{wang2020atloc}} & 1.53,\textbf{3.21} & 2.17,3.99 & 0.93,4.61 & 2.14,5.76 & 1.69,4.39 \\
\multicolumn{1}{l}{AtLoc w/ hypernet} & \textbf{0.96},3.43 & \textbf{2.04},\textbf{3.61} & \textbf{0.88},\textbf{4.22} & \textbf{1.72},\textbf{5.44} & \textbf{1.40,4.18}\\
\hline
\multicolumn{1}{l}{Baseline APR} & 0.89,2.29 & 1.49,3.32 & 0.74,4.79 & 1.40,4.95 & 1.13,3.84 \\
\multicolumn{1}{l}{Baseline APR w/ hypernet} & \textbf{0.77},\textbf{2.07} & \textbf{1.47},\textbf{3.29} & \textbf{0.72},\textbf{4.01} & \textbf{1.37},\textbf{4.90} & \textbf{1.08,3.57}\\ \hline
\end{tabular}
\vspace*{-1em}
\end{table*}

\textbf{Multi-Scene APR.} In contrast to \cite{ShavitFerensIccv21}, we
incorporated four-layer (instead of six) Transformers in each branch
(position and orientation) within the main network architecture. Each block
consists of an MHA layer with four heads, followed by an MLP that preserves
the input dimension. A dropout rate of $p=0.1$ was applied to each MLP
layer. The output dimension of each Transformer in the main network was $C_{M}^{Out}=256$. The position and orientation branches of the
hypernetwork generate the weights of three regression layers: input, hidden,
and output. The number of weights in the position and orientation branches
differ. The hypernetwork's position branch produces the weights for an MLP
with dimensions 256, 128, and 3, while the orientation branch generates
weights for an MLP with dimensions 512, 512, and 4. The weights generated by
the hypernetwork are then put into the regression heads of the main
network to estimate the position and orientation of the input query image.
The input and hidden layers of the regression head utilize the Swish
activation function.

\begin{table*}[tbh]
\caption{\textbf{Comparison to single-scene state-of-the-art methods - Cambridge Landmarks:} We report the median position/orientation error in meters/degrees. The most effective single and multi-scene APRs are marked in \textbf{bold}.}
\label{tb:cambridge_res_single_scene}
\centering{\
\small
\begin{tabular}{llccccc}
\toprule &  & \textbf{College} & \textbf{Hospital} & \textbf{Shop Facade} &
\textbf{St. Mary} & \textbf{Avg.} \\
\midrule & DSAC\textsuperscript{*}~\cite{9394752} & 0.18,0.3\degree & 0.21,0.4\degree & 0.05,0.3%
\degree & 0.15,0.5\degree & 0.15,0.4\degree \\
\midrule \multirow{2}{*}{{\rotatebox[origin=c]{90}{Seq.}}} & MapNet~\cite%
{brahmbhatt2018geometry} & 1.08,1.9\degree & 1.94,3.9\degree & 1.49,4.2%
\degree & 2.00,4.5\degree & 1.63,3.6\degree \\
& GL-Net~\cite{glnet} & 0.59,0.7\degree & 1.88,2.8\degree & 0.50,2.9\degree
& 1.90,3.3\degree & 1.22,2.4\degree \\
\midrule \multirow{2}{*}{{\rotatebox[origin=c]{90}{IR}}} & VLAD~\cite%
{denseVLAD} & 2.80,5.7\degree & 4.01,7.1\degree & 1.11,7.6\degree & 2.31,8.0%
\degree & 2.56,7.1\degree \\
& VLAD+Inter~\cite{sattler2019understanding} & 1.48,4.5\degree & 2.68,4.6%
\degree & 0.90,4.3\degree & 1.62,6.1\degree & 1.67,4.9\degree \\
\midrule \multirow{3}{*}{{\rotatebox[origin=c]{90}{RPR}}} & EssNet~\cite%
{essnet} & 0.76,1.9\degree & 1.39,2.8\degree & 0.84,4.3\degree & 1.32,4.7%
\degree & 1.08,3.4\degree \\
& NC-EssNet~\cite{essnet} & 0.61,1.6\degree & 0.95,2.7\degree & 0.7,3.4%
\degree & 1.12,3.6\degree & 0.85,2.8\degree \\
& RelocGNN\cite{glocker2013real} & 0.48,1.0\degree & 1.14,2.5\degree & 0.48,2.5%
\degree & 1.52,3.2\degree & 0.91,2.3\degree \\
\midrule \multirow{9}{*}{{\rotatebox[origin=c]{90}{APR}}} & PoseNet \cite%
{kendall2015posenet} & 1.92,5.40\degree & 2.31,5.38\degree & 1.46,8.08\degree
& 2.65,8.48\degree & \multicolumn{1}{l}{2.08,6.83\degree} \\
& BayesianPN \cite{kendall2016modelling} & 1.74,4.06\degree & 2.57,5.14%
\degree & 1.25,7.54\degree & 2.11,8.38\degree & \multicolumn{1}{l}{1.91,6.28%
\degree} \\
& LSTM-PN \cite{walch2017image} & 0.99,3.65\degree & {1.51},4.29\degree &
1.18,7.44\degree & 1.52,6.68\degree & \multicolumn{1}{l}{1.30,5.57\degree}
\\
& SVS-Pose \cite{naseer2017deep} & 1.06,2.81\degree & 1.50%
,4.03\degree & \textbf{0.63},5.73\degree & 2.11,8.11\degree &
\multicolumn{1}{l}{1.32,5.17\degree} \\
& GPoseNet \cite{cai2019hybrid} & 1.61,2.29\degree & 2.62,3.89\degree &
1.14,5.73\degree & 2.93,6.46\degree & \multicolumn{1}{l}{2.07,4.59\degree}
\\
& PoseNetLearn \cite{kendall2017geometric} & 0.99,{1.06%
\degree} & 2.17,\textbf{2.94\degree} & 1.05,3.97\degree & 1.49,%
3.43\degree & \multicolumn{1}{l}{1.42,2.85\degree} \\
& GeoPoseNet \cite{kendall2017geometric} & 0.88,\textbf{1.04}\degree
& 3.20,3.29\degree & 0.88,3.78\degree & 1.57,\textbf{3.32}\degree &
\multicolumn{1}{l}{1.63,2.86\degree} \\
& IRPNet \cite{shavitferensirpnet} & 1.18,2.19\degree & 1.87,3.38\degree &
0.72,\textbf{3.47\degree} & 1.87,4.94\degree & \multicolumn{1}{l}{
1.41,3.50\degree} \\
& \textbf{Baseline APR w/ hypernet} & \textbf{0.61},1.84\degree & \textbf{1.44},3.03\degree & 0.70,3.62\degree & \textbf{1.37},4.85\degree & \multicolumn{1}{l}{\textbf{1.03},\textbf{2.67\degree}} \\
\midrule \multirow{3}{*}{{\rotatebox[origin=c]{90}{MS-APR}}} & MSPN \cite{blanton2020extending} & 1.73,3.65\degree & 2.55,4.05\degree & 2.92,7.49\degree &
2.67,6.18\degree & \multicolumn{1}{l}{2.47,5.34\degree} \\
& MS-Trans\cite{shavit2021learning} & 0.83,1.47\degree & \textbf{1.81},2.39\degree & 0.86,\textbf{3.07\degree} & 1.62,3.99\degree & \multicolumn{1}{l}{1.28,2.73\degree} \\
& \textbf{MS-HyperPose} & \textbf{0.78},\textbf{1.18\degree} & 1.84,\textbf{2.11}\degree & \textbf{0.83},3.13\degree & \textbf{1.61},\textbf{3.22\degree} & \multicolumn{1}{l}{\textbf{1.27},\textbf{2.41\degree}} \\
\bottomrule &  &  &  &  &  &
\end{tabular}
}
\vspace*{-2em}
\end{table*}

%%%%%%%%%%%%%%%%%%%%%%%%%%%%%%%
%%% ECL Dataset
%%%%%%%%%%%%%%%%%%%%%%%%%%%%%%%

\section{Extended Cambridge Landmarks Dataset}
\label{appx_ecl_dataset}

\begin{figure}[!htbp]
    \centering
    \subfloat[Original]{\includegraphics[width=0.23\textwidth]{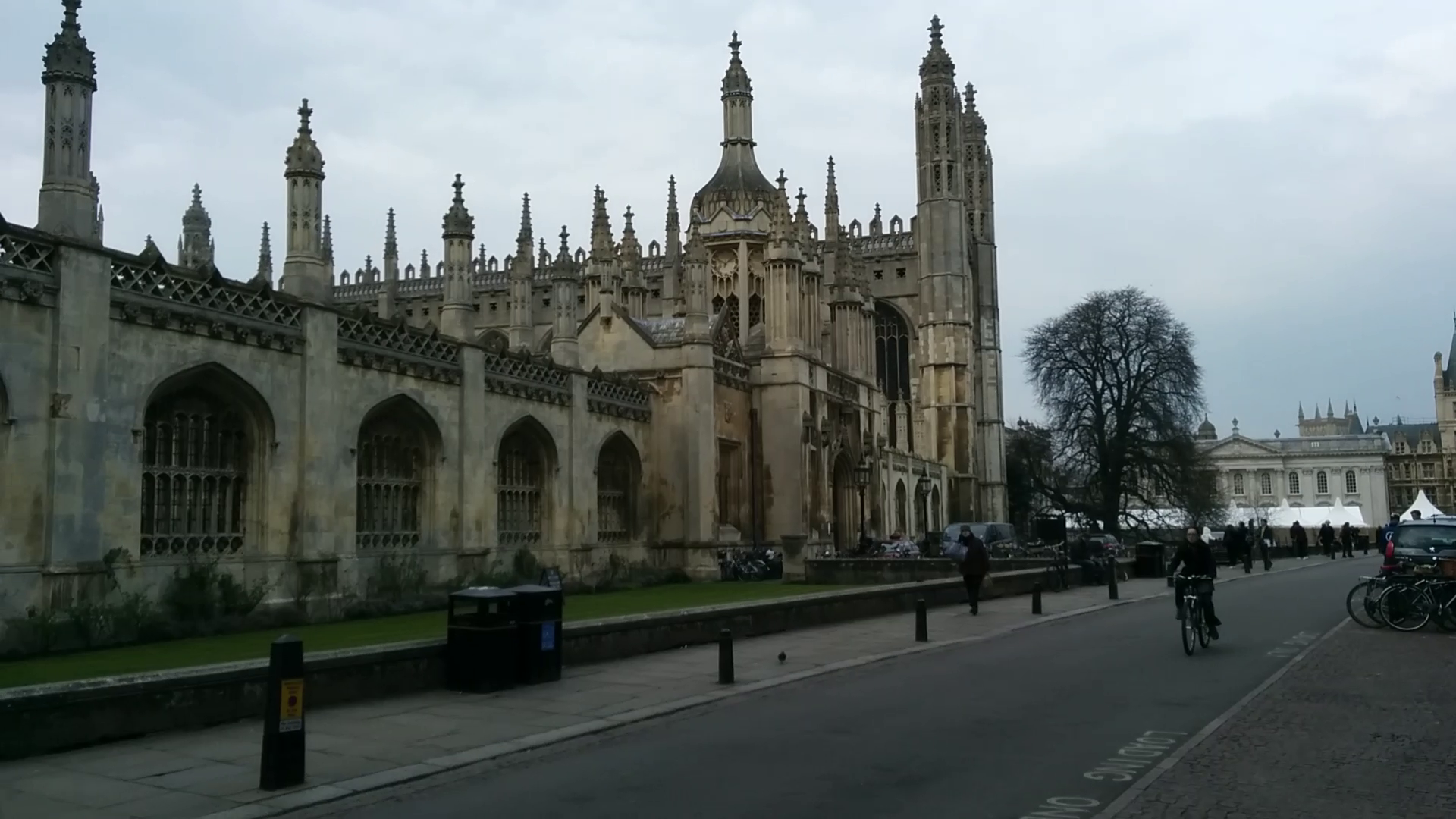}%
    }
    \hfill
    \subfloat[Evening]{\includegraphics[width=0.23\textwidth]{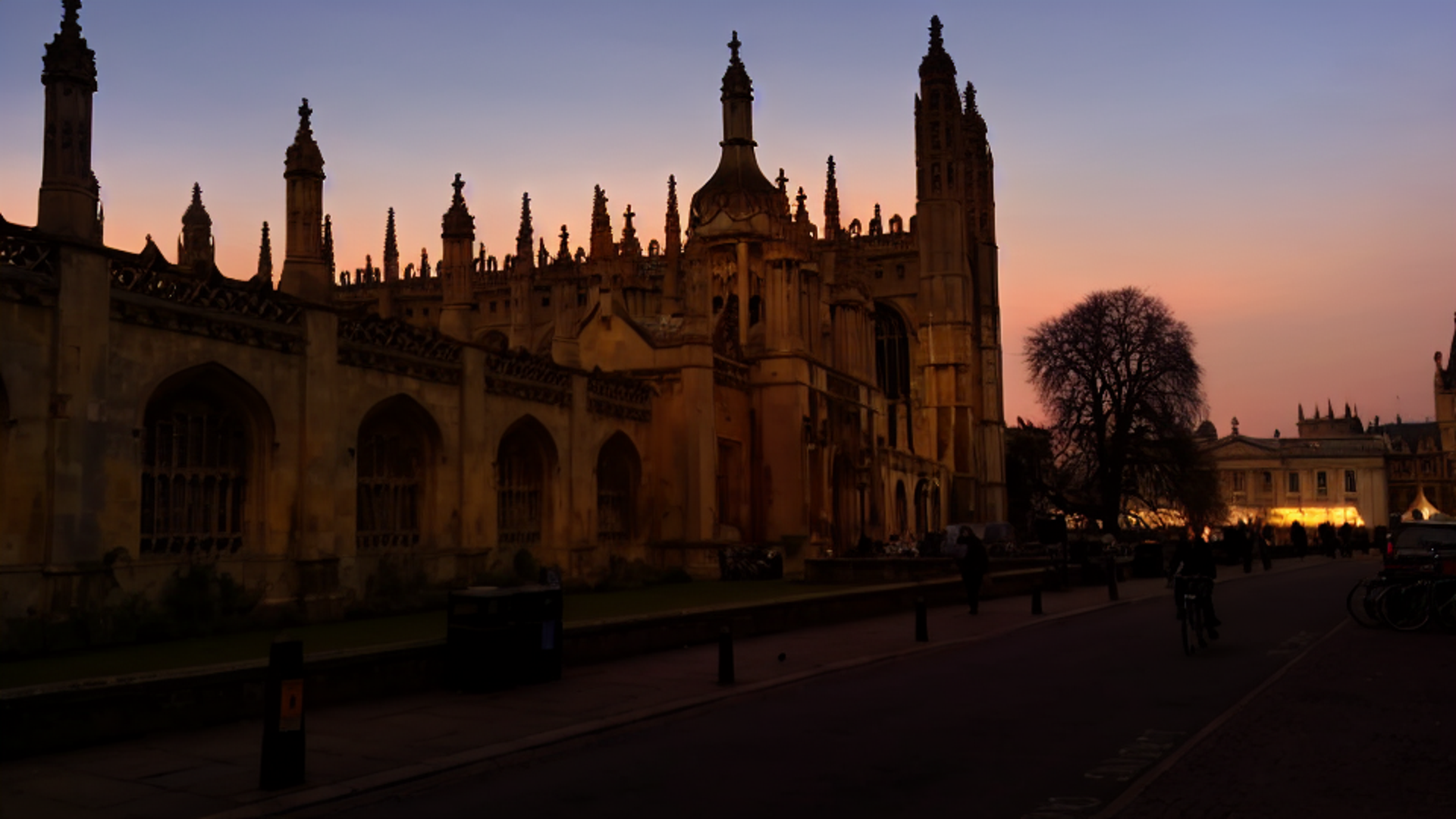}%
    }

    \subfloat[Summer]{\includegraphics[width=0.23\textwidth]{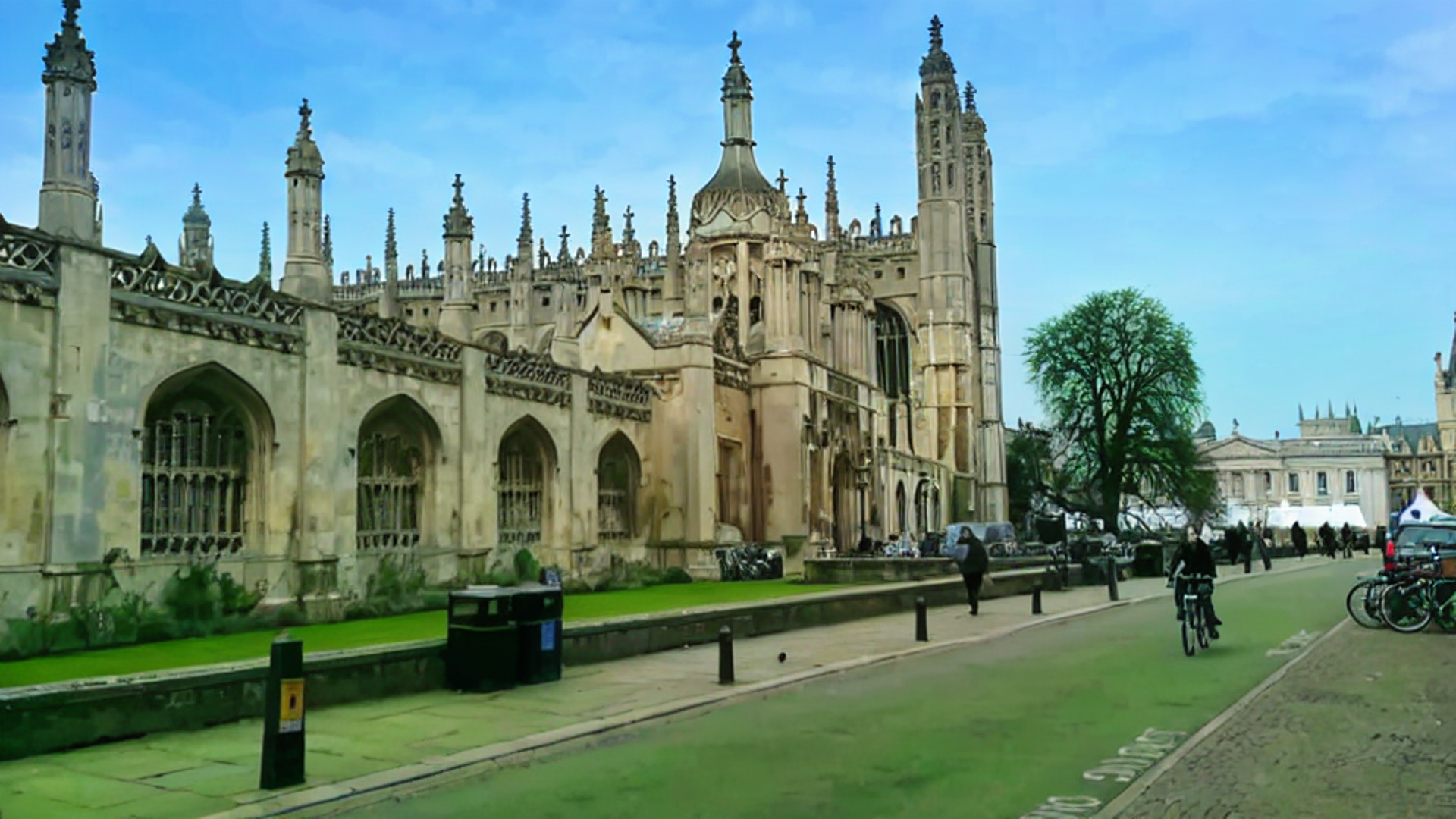}%
    }
    \hfill
    \subfloat[Winter]{\includegraphics[width=0.23\textwidth]{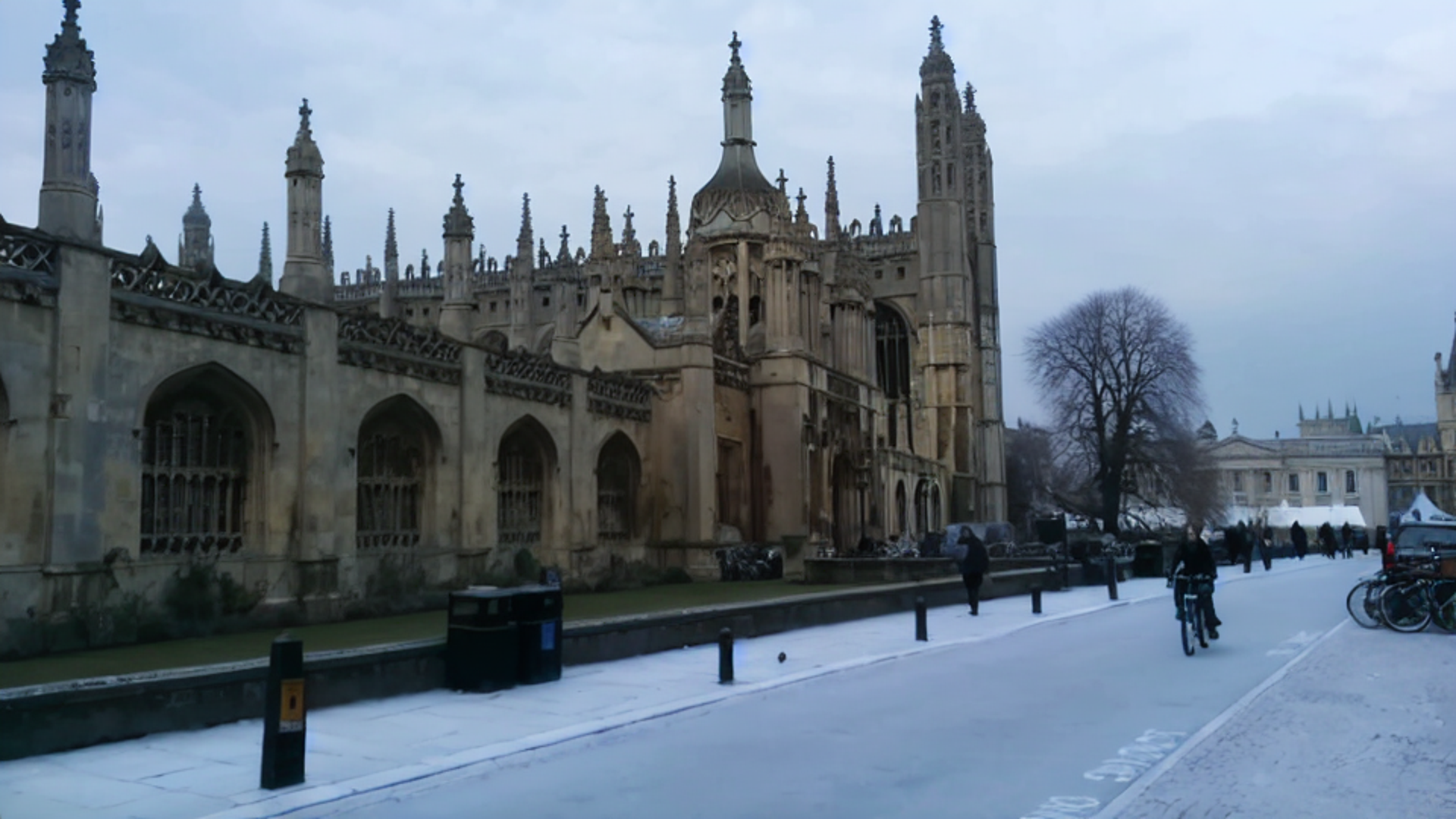}%
    }
    \caption{A sample from the King's College Scene in the proposed Extended Cambridge Landmarks (ECL) Dataset}
    \label{fig:ecl_dataset_kings_sample}
\end{figure}

To further assess the limitations of the HyperPose approach, we constructed an augmented version of the Cambridge Landmarks \cite{kendall2015posenet} benchmark called the Extended Cambridge Landmarks (ECL) dataset. Leveraging the original scenes, the ECL dataset introduces new localization challenges due to variations in appearance by creating synthetic evening, winter, and summer variants for each location. These scene permutations simulate real-world environmental variations by introducing shifts in lighting, weather patterns, foliage density, and more. Analyzing performance across ECL variations and their original Cambridge counterparts allows us to evaluate the robustness of models to real-world variations. These augmentations were conducted by utilizing the novel \textit{InstructPix2Pix} work \cite{brooks2022instructpix2pix}. Visual samples of the ECL dataset can be found in the supplementary materials. Figure \ref{fig:ecl_dataset_kings_sample} shows sample from the King's College scene in the extended dataset. Additional samples from each scene in the ECL dataset can be found in the supplementary material.
Quantitative pose regression analysis between these more challenging conditions and unmodified sequences helps scientifically assess model adaptability. We hypothesize that exposing limitations unseen in conventional test sets will motivate the development of more invariant solutions. As controllable expansions of established benchmarks, programmatically generated datasets can thus guide the development of robust models for real-world use.

%%%%%%%%%%%%%%%%%%%%%%%%%%%%%%%
%%% EXPERIMENTS
%%%%%%%%%%%%%%%%%%%%%%%%%%%%%%%

\section{Experimental Results}
\label{sec:experiments}
\vspace*{-0.3em}

We conducted an experimental evaluation using the 7 Scenes \cite{glocker2013real} and Cambridge Landmarks \cite{kendall2015posenet} datasets for evaluating absolute pose regression. The Cambridge Landmarks dataset represents an outdoor urban environment ranging in size from 900 to 5500 square meters. We present results for four of its six scenes. We excluded the remaining two scenes due to limited coverage in prior work. To compare robustness, we also report the results on the Extended Cambridge Landmarks dataset (ECL) (Section 1 in the supplementary materials). The 7 Scenes dataset consists of seven small-scale indoor scenes with spatial extents ranging from approximately 1 to 10 square meters. These datasets present various localization challenges, including differences in scale, indoor versus outdoor settings, repetitive elements, textureless features, significant viewpoint changes, and variations in trajectory between training and test sets. To facilitate a broader analysis of our proposed method, we conduct an additional evaluation on the Oxford RobotCar Dataset \cite{maddern20171}. Such dynamic environmental factors assess HyperPose's ability to improve localization accuracy amid diverse, challenging visual contexts. Our experimental procedures adhere to the same training and evaluation protocols as in \cite{brahmbhatt2018geometry,wang2020atloc}. Additional information concerning these sequences, and a comprehensive description of the training methodology and hyperparameter selection for the models, are provided in the supplementary material.

\begin{table*}[tbh]
\caption{\textbf{Comparison to single-scene SOTA methods - 7Scenes:} We report the median position/orientation error in meters/degrees. The most effective single and multi-scene APRs are distinguished in \textbf{bold}.}
\label{tb:7scenes_res_single_scene}
\centering
{
\small
\begin{tabular}{clcccccccc}
\toprule &  & \textbf{Chess} & \textbf{Fire} & \textbf{Heads} & \textbf{%
Office} & \textbf{Pumpkin} & \textbf{Kitchen} & \textbf{Stairs} & \textbf{Avg.} \\
\midrule \multirow{2}{*}{{\rotatebox[origin=c]{90}{SCR}}}
& DSAC\textsuperscript{*}~\cite{9394752} & 0.02,1.11\degree &
0.02,1.24\degree & 0.01,1.82\degree & 0.03,1.18\degree & 0.04,1.41\degree &
0.03,1.70\degree & 0.04,1.42\degree & 0.03,1.41\degree\\
& Marepo\cite{chen2024map} & 0.02,1.24\degree &
0.02,1.39\degree & 0.02,2.03\degree & 0.03,1.26\degree & 0.04,1.48\degree &
0.04,1.71\degree & 0.06,1.67\degree & 0.03,1.54\degree\\
\midrule \multirow{3}{*}{{\rotatebox[origin=c]{90}{Seq.}}}
& LsG~\cite{lsg}
& 0.09,3.3\degree & 0.26,10.9\degree & 0.17,12.7\degree & 0.18,5.5\degree &
0.20,3.7\degree & 0.23,4.9\degree & 0.23,11.3\degree & 0.19,7.74\degree\\
& MapNet\cite{brahmbhatt2018geometry} & 0.08,3.3\degree & 0.27,11.7\degree &
0.18,13.3\degree & 0.17,5.2\degree & 0.22,4.0\degree & 0.23,4.9\degree &
0.30,12.1\degree & 0.21,7.78\degree\\
& GL-Net\cite{glnet} & 0.08,2.8\degree & 0.26,8.9\degree & 0.17,11.4\degree
& 0.18,13.3\degree & 0.15,2.8\degree & 0.25,4.5\degree & 0.23,8.8\degree & 0.19,7.50\degree\\
\midrule \multirow{2}{*}{{\rotatebox[origin=c]{90}{IR}}}
& VLAD~\cite%
{denseVLAD} & 0.21,12.5\degree & 0.33,13.8\degree & 0.15,14.9\degree &
0.28,11.2\degree & 0.31,11.2\degree & 0.30,11.3\degree & 0.25,12.3\degree & 0.26,12.46\degree\\
& VLAD+Inter\cite{sattler2019understanding} & 0.18,10.0\degree & 0.33,12.4%
\degree & 0.14,14.3\degree & 0.25,10.1\degree & 0.26,9.4\degree & 0.27,11.1%
\degree & 0.24,14.7\degree & 0.24,11.71\degree\\
\midrule \multirow{5}{*}{{\rotatebox[origin=c]{90}{RPR}}}
& NN-Net~\cite{nn-net} &
0.13,6.5\degree & 0.26,12.7\degree & 0.14,12.3\degree & 0.21,7.4\degree &
0.24,6.4\degree & 0.24,8.0\degree & 0.27,11.8\degree & 0.21,9.30\degree \\
& RelocNet\cite{balntas2018relocnet} & 0.12,4.1\degree & 0.26,10.4\degree &
0.14,10.5\degree & 0.18,5.3\degree & 0.26,4.2\degree & 0.23,5.1\degree &
0.28,7.5\degree & 0.21,6.73\degree \\
& EssNet~\cite{essnet} & 0.13,5.1\degree & 0.27,10.1\degree & 0.15,9.9\degree
& 0.21,6.9\degree & 0.22,6.1\degree & 0.23,6.9\degree & 0.32,11.2\degree &
0.22,8.03\degree \\
& NC-EssNet~\cite{essnet} & 0.12,5.6\degree & 0.26,9.6\degree & 0.14,10.7%
\degree & 0.20,6.7\degree & 0.22,5.7\degree & 0.22,6.3\degree & 0.31,7.9%
\degree & 0.21,7.50\degree \\
& RelocGNN\cite{glocker2013real} & 0.08,2.7\degree & 0.21,7.5\degree & 0.13,8.70%
\degree & 0.15,4.1\degree & 0.15,3.5\degree & 0.19,3.7\degree & 0.22,6.5%
\degree & 0.16,5.24\degree \\
& {CamNet \cite{ding2019camnet}} & 0.04,1.7\degree & 0.03,1.7\degree &
0.05,2.00\degree & 0.04,1.60\degree & 0.04,1.6\degree & 0.04,1.6\degree &
0.04,1.5\degree & 0.04,1.70\degree \\
\midrule \multirow{10}{*}{{\rotatebox[origin=c]{90}{APR}}}
& PoseNet \cite{kendall2015posenet} & 0.32,8.12\degree & 0.47,14.4\degree & 0.29,12.0\degree & 0.48,7.68\degree & 0.47,8.42\degree & 0.59,8.64\degree & 0.47,13.8\degree & 0.44,10.4\degree \\
& BayesianPN \cite{kendall2016modelling} & 0.37,7.24\degree & 0.43,13.7%
\degree & 0.31,12.0\degree & 0.48,8.04\degree & 0.61,7.08\degree & 0.58,7.54%
\degree & 0.48,13.1\degree & 0.47,9.81\degree \\
& LSTM-PN \cite{walch2017image} & 0.24,5.77\degree & 0.34,11.9\degree &
0.21,13.7\degree & 0.30,8.08\degree & 0.33,7.00\degree & 0.37,8.83\degree &
0.40,13.7\degree & 0.31,9.85\degree \\
& GPoseNet \cite{cai2019hybrid} & 0.20,7.11\degree & 0.38,12.3\degree &
0.21,13.8\degree & 0.28,8.83\degree & 0.37,6.94\degree & 0.35,8.15\degree &
0.37,12.5\degree & 0.31,9.95\degree \\
& PoseNetLearn\cite{kendall2017geometric} & 0.14,4.50\degree & 0.27,11.8%
\degree & 0.18,12.1\degree & 0.20,5.77\degree & 0.25,4.82\degree & 0.24,5.52%
\degree & 0.37,10.6\degree & 0.24,7.87\degree \\
& GeoPoseNet\cite{kendall2017geometric} & 0.13,4.48\degree & 0.27,11.3\degree
& 0.17,13.0\degree & 0.19,5.55\degree & 0.26,4.75\degree & 0.23,\textbf{5.35}%
\degree & 0.35,12.4\degree & 0.23,8.12\degree \\
& IRPNet\cite{shavitferensirpnet} & 0.13,5.64\degree & \textbf{0.25},9.67\degree &
\textbf{0.15},13.1\degree & 0.24,6.33\degree & 0.22,5.78\degree & 0.30,7.29\degree &
0.34,11.6\degree & 0.23,8.49\degree \\
& AtLoc\cite{wang2020atloc} & 0.10,\textbf{4.07}\degree & \textbf{0.25},11.4\degree
& 0.16,11.8\degree & 0.17,5.34\degree & 0.21,\textbf{4.37}%
\degree & 0.23,5.42\degree & \textbf{0.26},10.5\degree & 0.20,7.56\degree \\
& TransBoNet\cite{song2024transbonet} & 0.11,4.48\degree & \textbf{0.25},12.5\degree
& 0.18,14.0\degree & 0.20,\textbf{5.08}\degree & 0.19,4.77%
\degree & 0.17,\textbf{5.35}\degree & 0.30,13.0\degree & 0.20,8.45\degree \\
& \textbf{\makecell[l]{Baseline APR \\ \quad w/ hypernet}} & \textbf{0.09},4.47\degree & 0.28,\textbf{9.44\degree} & \textbf{0.15},\textbf{11.4\degree} & \textbf{0.16},6.29\degree & %
\textbf{0.19},4.62\degree & \textbf{0.18},6.95\degree &
0.28,\textbf{9.15\degree} & \textbf{0.19,7.48}\degree \\
\midrule \multirow{3}{*}{{\rotatebox[origin=c]{90}{MS-APR}}}
& MSPN\cite{blanton2020extending} & \textbf{0.09},4.76\degree & 0.29,10.5\degree &
0.16,13.1\degree & \textbf{0.16},6.80\degree & {0.19},5.50\degree & {0.21},6.61\degree & 0.31,11.6\degree & 0.20,7.56\degree\\
& MS-Trans\cite{shavit2021learning} & 0.11,4.66\degree & 0.24,9.60\degree & 0.14,12.2\degree & 0.17,\textbf{5.66}\degree & 0.18,\textbf{4.44}\degree & \textbf{0.17},\textbf{5.94}\degree & \textbf{0.26},8.45\degree & 0.18,7.28\degree\\
& \textbf{MS-HyperPose} & 0.11,\textbf{4.34}\degree & \textbf{0.23},9.79\degree & \textbf{0.13},\textbf{10.7}\degree & 0.17,6.05\degree & \textbf{0.16},5.24\degree & \textbf{0.17},6.86\degree & 0.27,\textbf{6.00\degree} & \textbf{0.18,7.00\degree}\\
\bottomrule &  &  &  &  &  &  &  &
\end{tabular}
}
\vspace*{-1em}
\end{table*}

\begin{table*}[!tbh]
\caption{\textbf{Comparative analysis of different APR architectures with and
without the proposed hypernetwork to the Oxford RobotCar
dataset:} We report the mean and median position/orientation error in
meters/degrees.}
\label{table:robotcar_results}
\centering
\small
\begin{tabular}{@{}l|cc|cc|ccl@{}}
\toprule
& \multicolumn{2}{c}{\textbf{MapNet}\cite{brahmbhatt2018geometry}} & \multicolumn{2}{c}{\textbf{AtLoc}\cite{wang2020atloc}} & \multicolumn{2}{c}{\textbf{AtLoc w/ hypernet}} \\ \midrule
Sequence & Mean & Median & Mean & Median & Mean & Median \\ \midrule
LOOP1 & 8.76, 3.46 & 5.79, 1.54 & 8.61, 4.58 & 5.68, 2.23 & \textbf{8.42, 3.30} & \textbf{4.91, 1.33} \\
LOOP2 & 9.84, 3.96 & 4.91, 1.67 & 8.61, 4.58 & 5.68, 2.23 & \textbf{8.10, 3.04} & \textbf{4.73, 1.27} \\
FULL1 & 41.4, 12.5 & 17.94, 6.68 & 29.6, 12.4  & 11.10, 5.28 & \textbf{29.58, 6.77} & \textbf{7.86, 1.38} \\
FULL2 & 59.3, 14.8 & 20.04, 6.39 & 48.2, 11.1 & 12.2, 4.63 & \textbf{45.96, 9.60} & \textbf{8.38, 1.45} \\ \bottomrule
\end{tabular}
\vspace*{-1em}
\end{table*}

%=========================================
% Results
%=========================================
We evaluate the impact of integrating hypernetworks in single-scene APRs. To this end, we compare the accuracy of two model variations, with and without hypernetworks, and report median position and orientation errors. Table \ref{tb:singlescene_comapre_cambridge} presents results on the Cambridge dataset. The hypernetwork improves the accuracy of the proposed baseline APR and AtLoc \cite{wang2020atloc} models. The baseline APR model size is $26.7$MB with an inference time of $9.32$ms, increasing to $40.9$MB and $9.45$ms with a hypernetwork. The original published AtLoc model is $97.9$MB with an inference time of 5.14ms. The hypernetwork-enhanced AtLoc version has a model size of $116.8$MB and an inference time of $5.56$ms. Note that the reported errors refer to the models' performance after the first stage of model training, without applying the two additional optimization steps described in the training details (See supplementary materials). Furthermore, Tables \ref{tb:cambridge_res_single_scene} and \ref{tb:7scenes_res_single_scene} compare the median position error, and the orientation error, of the proposed baseline APR with hypernetwork architecture, with the results of recent SOTA localization algorithms, using the Cambridge Landmarks and 7Scenes datasets, respectively. For the indoor and outdoor datasets, in addition to the results of leading APRs, we report the median position and orientation errors of structure-based, sequence-based, IR, and RPR methods. Although DSAC\textsuperscript{*}~\cite{9394752} achieves SOTA accuracy, compared to all other APR methods, HyperPose obtains the most accurate average of $1.03m$, $2.67\degree$ on the Cambridge outdoor dataset, and $0.19m$, $7.48\degree$ on the 7Scenes indoor dataset. These localization errors show competitive performance compared to leading RPR techniques, with the hypernet-enhanced baseline ranking third and second/fourth (7Scenes/Cambridge) in positional and rotational accuracy, respectively.
Next, we compare the proposed MS-HyperPose architecture with other multiscene APR methods. Tables \ref{tb:cambridge_res_single_scene} and \ref{tb:7scenes_res_single_scene} present median position and orientation errors, respectively, on the Cambridge Landmarks and 7 Scenes datasets. On both indoor and outdoor benchmarks, our MS-HyperPose approach achieves the most accurate average of $1.27$m, $2.41\degree$ on Cambridge and $0.18$m, $7.12\degree$ on 7Scenes compared to other single- and multi-scene APRs. Additionally, MS-HyperPose is the most accurate for both positional and rotational errors in 6 out of 8 parameters on Cambridge and 6 out of 14 on 7 Scenes. The runtime of MS-HyperPose is $30.93$ms, similar to the reference architecture of \cite{shavit2021learning}. Its model size is $571$MB, notably larger than the $71.1$MB of the original. We discuss storage considerations in the \textit{Limitation and Future Work} section of the supplementary materials. 

Table \ref{table:robotcar_results} reports the mean and median position and
orientation errors of AtLoc \cite{wang2020atloc} compared to our hypernetwork-enhanced variant using the Oxford RobotCar Dataset \cite{maddern20171}. We also compare the enhanced model with \cite{brahmbhatt2018geometry}, the SOTA APR for this dataset using image sequences. Visualizations of the localization results can be found in the supplementary materials.

We also assess the performance of two configurations of the baseline APR model, with and without hypernetworks on the Extended Cambridge dataset (ECL). Across all scenes and variations, the hypernetwork-enhanced model outperforms the baseline, demonstrating better adaptation to the seasonal changes and diverse lighting conditions inherent in the scenes. The detailed analysis elucidating the robustness and resilience of HyperPose to scene changes is in the supplementary material.

%%%%%%%%%%%%%%%%%%%%%%%%%%%%%%%
%%% CONCLUSIONS
%%%%%%%%%%%%%%%%%%%%%%%%%%%%%%%
\section{Conclusions}
\label{sec:conclusions}
In this work, we proposed a hypernetwork-based approach to improve APR models. Our experiments show that hypernetworks improve accuracy for single-scene APR models. We also present MS-HyperPose, a novel multiscene camera pose regression model enhanced with a hypernetwork, which achieves new SOTA accuracy in indoor and outdoor localization datasets. We introduce the Extended Cambridge Landmarks (ECL) dataset to facilitate research on robust localization methods.

%%%%%%%%%%%%%%%%%%%%%%%%%%%%%%%
%%%%%%%%%%%%%%%%%%%%%%%%%%%%%%%
%%%%%%%%%%%%%%%%%%%%%%%%%%%%%%%
{
    \small
    \bibliographystyle{ieeenat_fullname}
    \bibliography{localization.bbl}
}

\end{document}